\documentclass{article}

\usepackage[preprint, nonatbib]{neurips_2020}
\usepackage[numbers]{natbib}
\usepackage[utf8]{inputenc} 
\usepackage[T1]{fontenc}    
\usepackage{hyperref}       
\usepackage{url}            
\usepackage{booktabs}       
\usepackage{amsfonts}       
\usepackage{nicefrac}       
\usepackage{microtype}      
\usepackage{amsmath}
\usepackage{graphicx}
\usepackage[normalem]{ulem} 
\usepackage{color}          
\usepackage{subfig}
\usepackage{textcomp}
\title{Graph Neural Network for Metal Organic Framework Potential Energy Approximation }
\author{%
  Shehtab Zaman \\
  Department of Computer Science\\
  State University of New York at Binghamton\\
  Binghamton, NY 13850 \\
  \texttt{szaman5@binghamton.edu} 
   \And
   Christopher Owen \\
   Department of Physics, Applied Physics, and Astronomy\\
  State University of New York at Binghamton\\
  Binghamton, NY 13850 \\
   \texttt{cowen1@binghamton.edu} \\
   \AND
   Kenneth Chiu \\
   Department of Computer Science\\
  State University of New York at Binghamton\\
  Binghamton, NY 13850 \\
   \texttt{kchiu@binghamton.edu} \\
   \And
   Micheal Lawler \\
   Department of Physics, Applied Physics, and Astronomy\\
  State University of New York at Binghamton\\
  Binghamton, NY 13850 \\
   \texttt{mlawler@binghamton.edu} \\
}
\date{September 2020}

\begin{document}

\maketitle

\begin{abstract}
Metal-organic frameworks (MOFs) are nanoporous compounds composed of metal ions and organic linkers.
MOFs play an important role in industrial applications such as gas separation, gas purification, and electrolytic catalysis.
Important MOF properties such a potential energy are currently computed via techniques such as density functional theory (DFT).
Although DFT provides accurate results, it is computationally costly.
We propose a machine learning approach for estimating the potential energy of candidate MOFs, decomposing it into separate pair-wise atomic interactions using a graph neural network.
Such a technique will allow high-throughput screening of candidates MOFs.
We also generate a database of 50,000 spatial configurations and high quality potential energy values using DFT.
\end{abstract}

\section{Introduction}


Metal-organic frameworks (MOFs) are a class of crystalline nanoporous materials composed of metal ions connected with organic ligands \cite{This_is_a_mof}. Due to the flexibility of combining hundreds of organic ligands with tens of  metal ions in thousands of network geometries, the configuration space of possible MOFs is massive \cite{colon2014high}. The large configuration space, highly tunable crystal structures, and nanometer sized pores make MOFs very promising for a variety of applications. Possible uses of MOFs include hydrogen storage, drug delivery systems, gas purification, and electrolytic catalysis \cite{Use_of_mofs}. Designing MOFs with desirable structural properties is a multi-billion-dollar challenge. 

The mechanical properties of MOFs can be tuned to produce desirable characteristics, so rapidly quantifying the properties is a key stage of any specific application. In principle, we can calculate properties for any materials or molecules using atomistic simulations \cite{sun2014learning}. In practice, the computational complexity increases $O(N^3)$ \cite{O(3)} as a function of system size, and due to the hundreds to thousands of atoms in a MOF's unit cell, significant approximations are necessary to make the simulations on hundreds of thousands of configurations feasible. 

One fundamental calculation needed to design MOFs for many applications is the potential energy of a given configuration of a MOF. For example, it can be used to compute the mechanical stability of the MOF. Currently, we must use costly DFT-based calculations\cite{inbook} to obtain the potential energy for a single atom configuration but a data-driven approach could speed up the  calculations and still infer meaningful structure-property relationships \cite{carleo2019machine}. Such an approach could extract the underlying force-fields\cite{forcefields} that govern the potential energy across the entire MOF configuration space. It can significantly enhance and alter the current computational techniques used to understand molecules and matter.

Recent advances in deep learning, especially graph neural networks, for materials science have enabled data-driven research on raw molecular data. We propose a graph convolution network with the graph representations of MOFs for interpretable predictions of MOF potential energies. The graph neural network (GNN) model approximates the potential energy function of the crystal as a neural-network series expansion. We use DFT calculations as ground truth and create a labeled dataset for 50k structural configurations.  

\section{Related Work}
Real-world data from chemistry, biology, and social sciences, are not easily represented with grid-like structures like images. Social networks, protein complexes, and molecules have a natural representation in graphs that capture the translational, rotational, and permutation symmetries present in the data. The field of geometric deep learning focuses on applying deep learning to non-euclidean domains such as graphs and manifolds~\cite{bronstein2017geometric}. Graph based message-passing neural networks, have enabled machine learning on different domains, especially quantum chemistry. Gilmer et al. developed a set of message passing networks to achieve high accuracy in predicting chemical properties of organic molecules \cite{gilmer2017neural}.Simonovksy and Komodakis extended graph neural networks to condition the graph convolution filters to condition on both node features and edge features for graph classification tasks \cite{simonovsky2017dynamic}. Xie and Grossman utilized graph convolutional networks to learn the properties of crystal structures. The CGNN is able to extract representations of crystal structures that are optimum for predicting targets calculated using DFT. Our work builds on the edge-conditioned graph convolutional networks with a modified convolution kernel.

\section{Model}

For a MOF molecule with N atoms, we wish to derive or identify a candidate potential function $U$. We wish to represent the potential of the entire molecule as a combination of arbitrary functions of neighboring nodes and their distances. Thus we have,

\begin{equation} \label{Potential}
    U = \sum_{i}^{N} \sum_{j \in \mathcal{N}_i} g_{i,j}(r(i,j))
\end{equation}

Where $\mathcal{N}_i$ are the neighbors of atom i. In our case, neighbor could be described as a bonded atom and r(i,j) is the distance between atoms i and j. Here we make the assumption that the each atom-pair, $i, j$ has a separate function $g_{i,j}$.
\subsection{GNN architecture}
The target of our model is the set of functions $g_{i,j}(r(i,j))$. We represent the crystal structures using graphs, such that each atom is represented by a node, and the edge represents the distance betweens the two atoms. 
We further assume that the bonds are not double counted. 
We can ensure that in a graphical representation by using directed edges and ensuring in edge list $\mathcal{E}$, we impose the condition: $ e_{ij} \in \mathcal{E} \rightarrow e_{ji} \notin \mathcal{E}$.

We begin with a dataset of Graphs $G_i$, and potential targets $y_i$. For a graph $G_i$ we have a set of nodes or atoms $x_n$, and an edge list $\mathcal{E}_i$. We can therefore define a neighborhood for each node $\mathcal{N}_{x_n}$, where the edges $e_{nj} = r(n,j)$.

We define a graph convolution operator, MOFGCN, similar to edge-conditioned convolutions described in \cite{gilmer2017neural, PhysRevLett.120.145301, simonovsky2017dynamic}, such that. 

\begin{equation}
    x^{t+1}_{n} = \sum_{X_m \in \mathcal{N}_{x^{t}_n}} h((x^{t}_n + x^{t}) \oplus e_{n,m})
\end{equation}

Here we set $h$ denotes a neural network, and $\oplus$ is the concatenation operator. One-hot encoding the node-features effectively allows the neural network to learn multiples functions. Our encoding allows us to have the same inputs for the same atomic interactions, therefore sharing the same weights across all similar atom pairs throughout the graph. 

We then define a global pooling operation on the final representation of the nodes, $x_n$. we define a simple over all the nodes in a graph, and also a graph attention based pooling, as seen in \cite{li2015gated}.

\begin{align}
    y_{pred} = \sum_{n} x^{l}_{n} &&
    y_{pred} =  \sum_{n} \sigma (h(x^{l}_{n})) \cdot j(x^{l}_{n})
\end{align}

where $\sigma$ is the softmax operator, and $h$ and $j$ are feed forward networks. 

For a given graph $G_i$, we can have a objective function that minimizes the distance between the target $y_i$ and the pooled quantity $R_i$. For the dataset we minimize the loss, 

\begin{equation} \label{objective}
\mathcal{L} = \frac{1}{N}\sum_{i} || y_i - y_{pred_i}||^2 
\end{equation}

We can see that if we minimize the Eq. \ref{objective}, we are able to retrieve neural network approximations of the functions,  $g_{i,j}(i,j,r(i,j))$ , from eq. \ref{Potential}. We use PyTorch and PyTorch Geometric to create and training the model \cite{Fey/Lenssen/2019, PyTorch}.

\begin{figure}
  \centering
  \includegraphics[width=11cm]{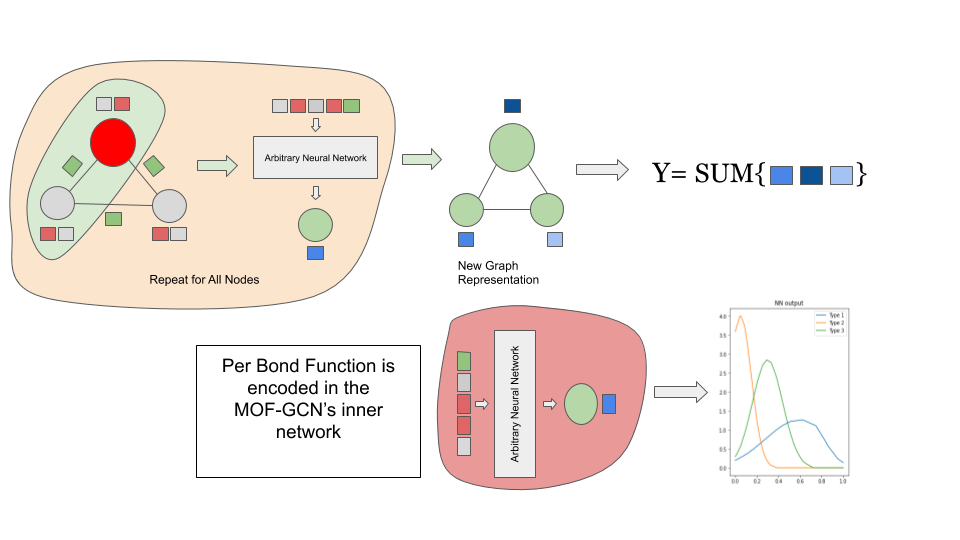}
  \caption{Schematic of Graph Neural Network combing Node and edge features. The convolutions on a 3 node graph with subsequent feature aggregation is shown. The convolution kernel operates on neighboring nodes with a neural network shared for all node-pairs. The SUM reduction of node features is shown.}
\end{figure}

\section{Experiments and Results}

\subsection{Proof-of-Concept}
\begin{figure}
    \centering
    \subfloat[\centering The NN is only trained with the real-combined targets, and, predicts the graph-level target value. We are able to extract the predicted functions the NN.  It's important  approximations are only correct up to an additive constant. \label{fig:3_func}]{\includegraphics[width=\linewidth]{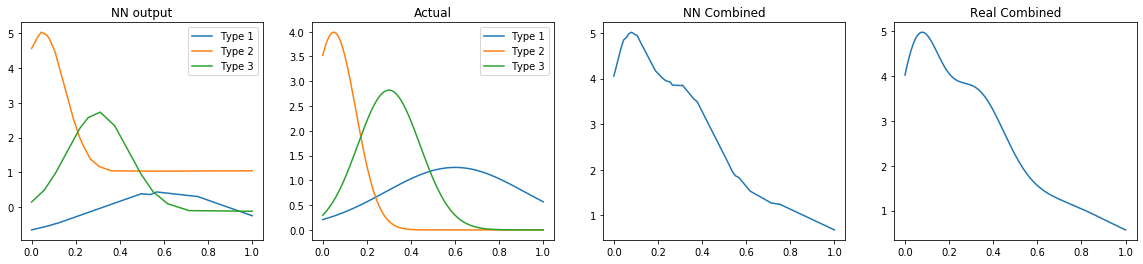}}
    
    \qquad
    \hspace{-1cm}
    \subfloat[\centering Histogram of test errors. Most predictions are within 10 Ry of the DFT calculated energies. \label{fig:test_errors}]{\includegraphics[width=0.32\linewidth]{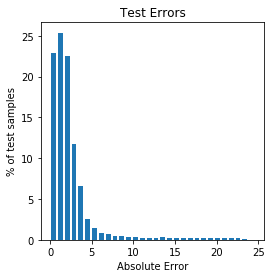}}
    \qquad
    \hspace{-.8cm}
    \subfloat[\centering Approximate interactions functions learned by the graph filter. Filter outputs are combined to predict the potential function.  \label{fig:atom_approximates}]{\includegraphics[width=0.66\linewidth]{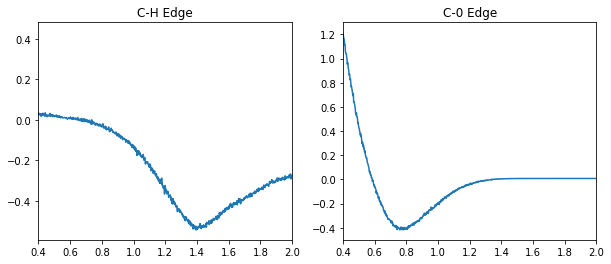}}
\end{figure}

We first sought to demonstrate that our approach could find a decomposition of known functions. We generate 10,000 three node graphs, with three distinct node types. The nodes are spaced apart by a random distance. The "energy" is for each node-pair is calculated with Gaussian probability functions with $\mu = [0.6,0.05,0.3]$ and $\sigma=[0.1,0.01,0.02]$. The graph target is a sum of the three "energies". We train the MOFGCN model to predict the graph-level target, and approximate the node-pair functions as seen in Fig. \ref{fig:3_func}.

\subsection{MOF Dataset}
This dataset is constructed using Quantum Espresso \cite{QE2009}, an ab initio software for electronic structure and energy calculations. We used the FIGXAU from the CoRE MOF database  \cite{chung2019advances}. We performed the structural optimization with the PWscf package \cite{PWSCF}. We found the ground state configuration using the Kjpaw \cite{PAW} pseudopotentials and the Perdew-Burke-Ernzerhof(PBE) \cite{PBE} exchange-correlation functional. From this ground state configuration, random fluctuations were introduced by allowing each atom to randomly move any rational number between $\pm5$(Å) either on its x,y or z axis. 47,617 new atomic configurations were generated and a Self-Consistent Field Calculation (SCF) was done for each one.  


We use the dataset to train the MOFGCN model with an attention based reduction to predict the potential energy. Each sample of the MOF is represented as a graph with the nodes being the atoms and the edge feature being the inter-atomic distance. We achieve results comparable results to DFT (Fig. \ref{fig:test_errors}) and are also able to estimate atomic interactions as seen in Fig. \ref{fig:atom_approximates}.
\section{Conclusion and Future Work}

The MOFGCN model learns effective functions that governs the potential energy of the MOF. The model achieves comparable accuracies to DFT at a fraction of the computation costs. The MOFGCN graph kernel produces an interpretable intermediate representation of the molecular graph inputs. We utilize the flexibility of neural networks to approximate arbitrary smooth functions to decompose complex interactions in a crystal lattice.We plan on further extending our dataset larger MOFs and expanding the number of atom-atom interactions learned by our model and enable rapid characterizations of MOFs.

Automatic discovery of scientific laws and principles using data-driven machine learning is a potentially transformational development in science.~\cite{zhang2020deep, butler2018machine, schmidt2009distilling}.
Our preliminary work here demonstrates that decomposition of the potential energy into the sum of functions is possible. Our future work will seek to demonstrate that these functions also have a physical, scientific significance.



\section*{Broader Impact}

We acknowledge that our work may have significant impact on issues relating to energy storage, carbon sequestration, and drug discovery. Hydrogen may play a key role in developing cleaner sources of energy.  Clean, renewable energy has a significant social impact.Gas storage, especially carbon dioxide, is also a significant ethical drive to further understand MOFs. Carbon sequestration is an important tool to mitigate the effects of climate change. Furthermore, the tuning MOFs for drug-delivery systems may also pose significant ethical issues both good and bad. 
\bibliography{bib}
\end{document}